\pdfoutput=1 
\documentclass[runningheads]{llncs}

\usepackage[ruled,linesnumbered]{algorithm2e}
\usepackage{svg}
\usepackage{amsmath,graphicx,bbm,bm, subfigure}
\usepackage{float, cite} 
\usepackage{subfigure}
\usepackage{booktabs} 
\usepackage[export]{adjustbox} 
\usepackage{amsfonts}

\newcommand{\mb}[1]{\mathbf{#1}}

\newcommand{\mbmu}{\bm{\mu}}
\newcommand{\mbtheta}{\bm{\theta}}
\newcommand{\mbSigma}{\mathbf{\Sigma}}
\newcommand{\mbOmega}{\bm{\Omega}}

\newcommand{\calN}{\mathcal{N}}

\begin{document}
\title{Adaptive unsupervised learning with enhanced feature representation for intra-tumor partitioning and survival prediction for glioblastoma}


%
\titlerunning{Adaptive learning with enhanced representation for glioblastoma}
%
\author{Yifan Li\inst{1} \and
Chao Li\thanks{Equal contribution.}\inst{2}\and
Yiran Wei\inst{2}\and
Stephen Price\inst{2}\and
Carola-Bibiane Schönlieb\inst{3} \and
Xi Chen\inst{1,4}}

\authorrunning{Yifan Li et al.}
\institute{Department of Computer Science, University of Bath, Bath, UK. \and \{Division of Neurosurgery, Department of Clinical Neurosciences, \and
Department of Applied Mathematics and Theoretical Physics, \and
Department of Physics\}, University of Cambridge, Cambridge, UK.\\
\email{yl3548@bath.ac.uk, cl647@cam.ac.uk, yw500@cam.ac.uk, sjp58@cam.ac.uk, 
cbs31@cam.ac.uk, xc841@bath.ac.uk/xc253@mrao.cam.ac.uk}} 

\maketitle           
\begin{abstract}
Glioblastoma is profoundly heterogeneous in regional microstructure and vasculature. Characterizing the spatial heterogeneity of glioblastoma could lead to more precise treatment. With unsupervised learning techniques, glioblastoma MRI-derived radiomic features have been widely utilized for tumor sub-region segmentation and survival prediction. However, the reliability of algorithm outcomes is often challenged by both ambiguous intermediate process and instability introduced by the randomness of clustering algorithms, especially for data from heterogeneous patients. 

In this paper, we propose an adaptive unsupervised learning approach for efficient MRI intra-tumor partitioning and glioblastoma survival prediction. A novel and problem-specific Feature-enhanced Auto-Encoder (FAE) is developed to enhance the representation of pairwise clinical modalities and therefore improve clustering stability of unsupervised learning algorithms such as K-means. Moreover, the entire process is modelled by the Bayesian optimization (BO) technique with a custom loss function that the hyper-parameters can be adaptively optimized in a reasonably few steps. The results demonstrate that the proposed approach can produce robust and clinically relevant MRI sub-regions and statistically significant survival predictions. 


\end{abstract}

\section{Introduction}
\label{sec:intro}

Glioblastoma is one of the most aggressive adult brain tumors characterized by heterogeneous tissue microstructure and vasculature. Previous research has shown that multiple sub-regions (also known as tumor habitats) co-exist within the tumor, which gives rise to the disparities in tumor composition among patients and may lead to different patient treatment response~\cite{li2019intratumoral,li2019low}. Therefore, this intra-tumoral heterogeneity has significantly challenged the precise treatment of patients. Clinicians desire a more accurate identification of intra-tumoral invasive sub-regions for targeted therapy.

Magnetic resonance imaging (MRI) is a non-invasive technique for tumor diagnosis and monitoring. MRI radiomic features~\cite{sala2017unravelling} provide quantitative information for both tumor partition and survival prediction~\cite{li2019decoding,li2019multi}. Mounting evidence supports the usefulness of the radiomic approach in tumor characterization, evidenced by the Brain Tumor Image Segmentation (BraTS) challenge, which provides a large dataset of structural MRI sequences, i.e., T1-weighted, T2-weighted, post-contrast T1-weighted (T1C), and fluid attenuation inversion recovery (FLAIR). Although providing high tissue contrast, these weighted MRI sequences are limited by their non-specificity in reflecting tumor biology, 
where physiological MRIs, e.g., perfusion MRI (pMRI) and diffusion MRI (dMRI), could complement. Specifically, pMRI measures vascularity within the tumor, while dMRI estimates the brain tissue microstructure. Incorporating these complementary multi-modal MRI has emerged as a promising approach for more accurate tumor characterization and sub-region segmentation for clinical decision support.
    
Unsupervised learning methods have been widely leveraged to identify the intra-tumoral sub-regions based on multi-modal MRI~\cite{wu2017unsupervised, syed2020multiparametric, patel2020clustering, dextraze2017spatial, park2021spatiotemporal,xia2018radiogenomics}. Standard unsupervised learning methods, e.g., K-means, require a pre-defined class number, which lacks concrete determination criteria, affecting the robustness of sub-region identification. For instance, some researchers used pre-defined class numbers according to empirical experience before clustering~\cite{dextraze2017spatial, park2021spatiotemporal}. Some other work~\cite{xia2018radiogenomics,meyer2007unsupervised} introduced clustering metrics, e.g., the Calinski-Harabasz (CH) index, which quantifies the quality of clustering outcomes to estimate the ideal class number. However, the CH index is sensitive to data scale~\cite{xia2018radiogenomics, meyer2007unsupervised}, limiting its generalization ability across datasets. Some other clustering techniques, e.g., agglomerative clustering, do not require a pre-defined class number and instead require manual classification. A sensitivity hyper-parameter, however, is often needed \textit{a priori}. The clustering results can be unstable during iterations and across datasets. Due to the above limitations, the generalization ability of clustering methods has been a significant challenge in clinical applications, particularly when dealing with heterogeneous clinical data.

Further, the relevance of clustering results is often assessed using patient survival in clinical studies~\cite{leone2019overall, mangla2014correlation, beig2020radiogenomic, park2021spatiotemporal}. However, existing research seldom addressed the potential influence of instability posed by the unsupervised clustering algorithms. Joint hyper-parameter optimization considering both clustering stability and survival relevance is desirable in tumor sub-region partitioning.



In this paper, we propose a variant of auto-encoder (AE), termed Feature-enhanced Auto-Encoder (FAE), to identify robust latent feature space constituted by the multiple input MRI modalities and thus alleviate the impact brought by the heterogeneous clinical data. Additionally, we present a Bayesian optimization (BO) framework~\cite{snoek2012practical} to undertake the joint optimization task in conjunction with a tailored loss function, which ensures clinical relevance while boosting clustering stability. As a non-parametric optimization technique based on Bayes' Theorem and Gaussian Processes (GP)~\cite{rasmussen2003gaussian}, BO learns the representation of the underlying data distribution that the most probable candidate of the hyper-parameters is generated for evaluation in each step. Here, BO is leveraged to identify the (sub)optimal hyper-parameter set with the potential to effectively identify robust and clinically relevant tumor sub-regions. The primary contributions of this work include: 
\begin{itemize}
     \item Developing a novel loss function that balances the stability of sub-region segmentation and the performance of survival prediction.
     
    \item Developing an FAE architecture in the context of glioblastoma studies to further enhance individual clinical relevance between input clinical features and improve the robustness of clustering algorithms.

    \item Integrating a BO framework that enables automatic hyper-parameter search, which significantly reduces the computational cost and provides robust and clinically relevant results.
\end{itemize}

The remainder of this paper is organized as follows. Section 2 describes the overall study design, the proposed framework, and techniques. Section 3 reports numerical results, and Section 4 is the concluding remarks.

\section{Problem formulation and methodology}
\label{sec:format}

Consider an $N$ patients multi-modal MRI dataset $\mbOmega$ with $M$ modalities defined as $\{\mb{X}_m\}_{m=1}^M$. $\mb{X}_{m}$ denotes the $m$th (pixel-wise) modality values over a collection of $N$ patients. $\mb{X}_{m} = \{ \mb{x}_{m,n} \}_{n=1}^{N}$, where $\mb{x}_{m,n} \in {\mathbb{R}}^{I_{m,n}\times 1}$ and $I_{m,n}$ denotes total pixel number of an individual MRI image for the $m$th modality of the $n$th patient. 

Our goal is to conduct sub-region segmentation on MRI images and perform clinically explainable survival analysis. Instead of running unsupervised learning algorithms directly on $\mb{X}_m$, we introduce an extra latent feature enhancement scheme (termed FAE) prior to the unsupervised learning step to further improve the efficiency and robustness of clustering algorithms.


 
\begin{figure}[!ht]
    \centering
        \includegraphics[width=1\textwidth]{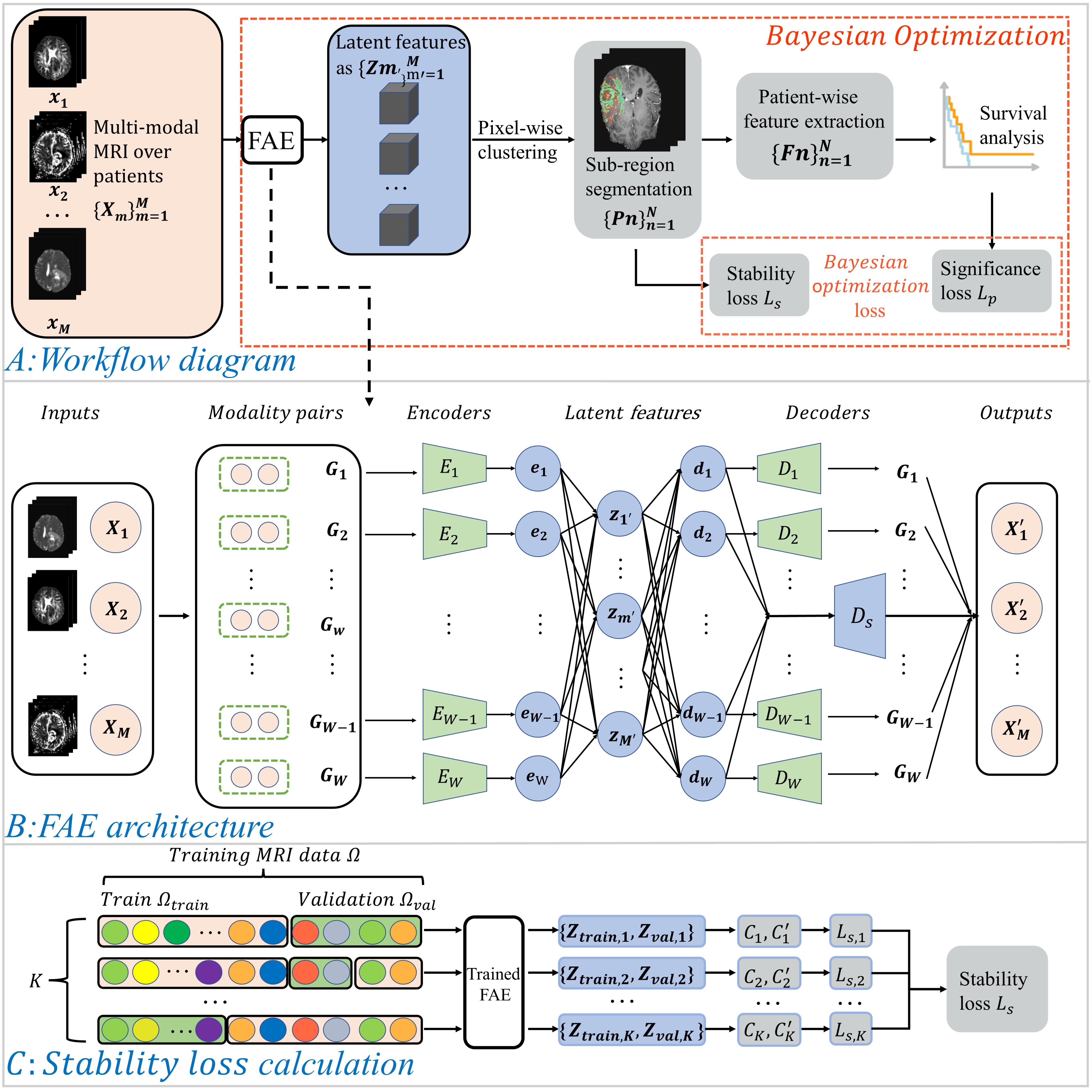}
    \caption{A: Workflow of the proposed approach. The entire process is modelled under a Bayesian optimization framework. B: Architecture of FAE. The light orange circle represents modality $\mb{X}_m$ overall patients and the blue circle is the latent feature $\mb{Z}_{m^{\prime}}$. The green dotted frame denotes the modality pair, and the green trapezoid represents feature-enhanced encoder $E$ and decoder $D$. The blue trapezoid indicates the fully connected decoders $D_s$. C: Illustration of stability loss calculation. Circles in different colours represent individual patient MRI data, which are then randomly shuffled for $K$ times to split into train/validation sets.}
\label{fig: flowchart}
\vspace{-0.4cm}
\end{figure}

As shown in Figure~\ref{fig: flowchart}(A), FAE aims to produce a set of latent features $\{\mb{Z}_{m^{\prime}}\}_{m^{\prime}=1}^M$ that represent the original data $\{\mb{X}_m\}_{m=1}^M$. Unlike a standard AE that takes all modalities as input, FAE `highlights' pairwise common features and produces $\mb{Z}$ through a set of encoders (denoted as $E$) and decoders (denoted as $D$). The latent features are then used in unsupervised clustering to classify tumor sub-region $\{ \mb{P}_n \}_{n=1}^N$ for all patients. As an intermediate step, we can now produce spatial features $\{\mb{F}_n \}_{n=1}^N$ from the segmented figures through radiomic spatial feature extraction methods such as gray level co-occurrence matrix (GLCM) and Gray Level Run Length Matrix (GLRLM)~\cite{mohanty2011classifying}. 

\subsection{Feature-enhanced auto-encoder}
FAE is developed on Auto-encoder (AE), a type of artificial neural network used for dimensionality reduction. A standard AE is a 3-layer symmetric network that has the same inputs and outputs. 
As illustrated in Figure~\ref{fig: flowchart}(B), FAE contains $W$ feature-enhanced encoder layers $\{E_w\}_{w=1}^W$ to deal with $\{\mb{G}_w\}_{w=1}^W$ pairs of modalities, where $W = \binom{M}{2}$ pairs of modalities (from combination) given $M$ inputs. The $w$th encoder takes a pair of modalities from $\{\mb{X}_m\}_{m=1}^M$ and encodes to a representation $\mb{e}_w$. The central hidden layer of FAE contains $\{\mb{Z}_{m^{\prime}}\}_{m^{\prime}=1}^M$ nodes that represents $M$ learnt abstract features. FAE also possesses a `mirrored' architecture similar to AE, where $W$ feature-enhanced decoder layers $\{D_w\}_{w=1}^W$ are connected to the decoded representations $\{d_w\}_{w=1}^W$. 

Unlike the standard symmetric AE, FAE has a `dual decoding' architecture that an extra fully-connected decoder layer $D_s$ is added to the decoding half of the networks to connect $\{d_w\}_{w=1}^W$ directly to the outputs $\{\mb{X^{\prime}}_m\}_{m=1}^M$. Decoder $D_s$ aims to pass all outputs information (and correlations) rather than the pairwise information from $\mb{G}_w$ in the back-propagation process. As a result, node weights 
$\{\mb{Z}_{m^{\prime}}\}_{m^{\prime}=1}^M$ are updated by gradients from both $\{D_w\}_{w=1}^W$ and $D_s$. In practice, $\mb{Z}$ and the encoders are iteratively amended by $\{D_w\}_{w=1}^W$ (i.e., reconstruction loss from pairwise AEs) and $D_s$ (i.e., global reconstruction loss) in turns. 

FAE enhances the latent features in every pair of input modalities before reducing the dimensionality from $W$ to $M$. For instance, $\mb{e}_w$ is a unique representation that only depends on (and thus enhances the information of) the given input pair $\mb{G}_w$. Under this dual decoding architecture, FAE takes advantage of highlighting the pairwise information in $\{\mb{Z}_{m^{\prime}}\}_{m^{\prime}=1}^M$ while retaining the global correlation information from $D_s$. Another advantage of FAE lies in its flexibility to the dimensionality of input features. The FAE presented in this paper always produces the same number of latent features as the input dimension. The latent dimension might be further reduced manually depending on computational/clinical needs.

\subsection{Patient-wise feature extraction and survival analysis}

We implement Kaplan–Meier (KM) survival analysis~\cite{beig2020radiogenomic,park2021spatiotemporal} on spatial features and sub-region counts $\{ \mb{F}_n \}_{n=1}^N$ to verify the relevance of clustering sub-regions. To characterize the intratumoral co-existing sub-regions, we employed the commonly used texture features from the GLCM and GLRLM families, i.e., Long Run Emphasis (LRE), Relative mutual information (RMI), Joint Energy, Run Variance (RV) and Non-Uniformity. These features are formulated to reflect the spatial heterogeneity of tumor sub-regions. For example, LRE indicates the prevalence of a large population of tumor sub-regions. The formulas and interpretations of all these features are detailed in~\cite{van2017computational}. We next use the k-medoids technique to classify $N$ patients into high- and low-risk subgroups based on $\{ \mb{F}_n \}_{n=1}^N$ and then perform KM  analysis to analyze the survival significance of the subgroups to determine the $L_p$, as described in Section 2.4 and equation~\ref{Eq:ploss}.

\subsection{Constructing problem-specific losses}
\subsubsection{Stability loss}
\label{ssec:subhead}
We first introduce a stability quantification scheme to evaluate clustering stability using pairwise cluster distance~\cite{von2010clustering,meilua2003comparing}, which will serve as part of the loss function in hyper-parameter optimization. Specifically, we employ a Hamming distance method (see~\cite{von2010clustering} for details) to quantify the gap between clustering models. We first split the MRI training dataset $\mbOmega$ into train and validation sets, denoted as $\mbOmega_{train}$ and $\mbOmega_{val}$ respectively. We then train two clustering models $C$ (based on $\mbOmega_{train}$) and $C'$ (based on $\mbOmega_{val}$). The stability loss aims to measure the performance of model $C$ on the unseen validation set $\mbOmega_{val}$. The distance $d(\cdot)$ (also termed as $L_s$) is defined as:
\begin{align}
    L_s = d(C,C')=\min_{\pi}\frac{1}{I_{val}}\sum_{\mbOmega_{val}} \mathbbm{1}_{\{\pi(C(\mbOmega_{val})\neq C'(\mbOmega_{val}))\}},
    \label{Eq:stability}
\end{align}
where $I_{val}$ denotes the total number of pixels over all MRI images in the validation set $\mbOmega_{val}$. $\mathbbm{1}$ represents the Dirac delta function~\cite{zhang2020dirac} that returns 1 when the inequality condition is satisfied and 0 otherwise, and function $\pi(\cdot)$ denotes the repeated permutations of dataset $\mbOmega$ to guarantee the generalization of the stability measure~\cite{von2010clustering}. 

Figure~\ref{fig: flowchart} (C) shows the diagram for $L_s$ calculation, where $N$ patients are randomly shuffled for $K$ times to mitigate the effect of randomness. $K$ pairs of intermediate latent features $\{\mb{Z}_{train, k}, \mb{Z}_{val, k}\}_{k=1}^K$ are generated through FAE for training the clustering models $C$ and $C'$. We then compute $L_s$ over $K$ repeated trials. $L_s$ is normalized to range $[0,1]$, and smaller values indicates more stable clusterings. 



\subsubsection{Significance loss}

We integrate prior knowledge from clinical survival analysis and develop a significance loss $L_p$ to quantify clinical relevance between the clustering outcomes and patient survival, as demonstrated in the below equation:

\begin{equation}
L_p=\left\{
\begin{array}{lcl}
\frac{1}{1-\tau} \log(\frac{\tau}{p}) & &   {0 < p \leq \tau}\\
\\
-\log_{\tau}(\frac{\tau}{p}) & &   {\tau < p < 1}\\
\end{array} \right.
\label{Eq:ploss}
\end{equation}
where p represents p-value (i.e., statistical significance measure) of the log-rank test in the survival analysis and $\tau$ is a predefined threshold. 

This follows the clinical practice that a lower p-value implies that the segmented tumor sub-regions can provide sensible differentiation for patient survival. In particular, for $p$ less than the threshold, the loss equation returns a positive reward. Otherwise, for $p$ greater than or equal to $\tau$, the segmented tumor sub-regions are considered undesirable and the penalty increases linearly with $p$.


\subsection{Bayesian optimization}
Hyper-parameters tuning is computational expensive and often requires expert knowledge, both of which raise practical difficulties in clinical applications. In this paper, we consider two undetermined hyper-parameters: a quantile threshold $\gamma \in [0,1]$ that distinguishes outlier data points from the majority and cluster number $\eta$ for the pixel-wise clustering algorithm. We treat the entire process of Figure~\ref{fig: flowchart}(A) as a \textit{black-box system}, of which the input is the hyper-parameter set $\mbtheta = [\gamma, \eta]$ and the output is a joint loss $\mathcal{L}$ defined as: 

\begin{equation}
     \mathcal{L} = \alpha L_s + (1-\alpha ) L_p
\label{loss}
\end{equation}
where $\alpha$ is a coefficient that balances $L_s$ and $L_p$ and ranges between $[0, 1]$. 

\begin{algorithm}[!ht]
\caption{Bayesian optimization for hyper-parameter tuning}
\label{Alg2}
Initialization of GP surrogate $f$ and the RBF kernel $\mathcal{K}(\cdot)$\\
\While {not converged}{Fit GP surrogate model $f$ with $\{ \theta_j,\mathcal{L}_j \}_{j=1}^J$ \\
Propose a most probable candidate $\theta_{j+1}$ through Equation~\eqref{EI} \\
Run $ \bf Algorithm$~\ref{Alg1} with $\theta_{j+1}$, and compute loss $\mathcal{L}_{j+1}$\\
Estimate current optimal $\theta_{j+2}$ of the constructed GP surrogate $f^{\prime}$\\
Run $\bf Algorithm$~\ref{Alg1} with $\theta_{j+2}$, calculate the loss $\mathcal{L}_{j+2}$\\
$J=J+2$\\}
Obtain (sub)optimal $\theta_{\ast}$ upon convergence 
\end{algorithm}

We address the hyper-parameter tuning issue by modelling the black-box system under BO, a sequential optimization technique that aims to approximate the search space contour of $\mbtheta$ by constructing a Gaussian Process (GP) surrogate function in light of data. BO adopts an \textit{exploration-exploitation scheme} to search for the most probable $\mbtheta$ candidate and therefore minimize the surrogate function mapping $f: \Theta \rightarrow \mathcal{L}$ in $J$ optimization steps, where $\Theta$ and $\mathcal{L}$ denote input and output spaces respectively. The GP surrogate is defined as: $f \sim \mathcal{GP}(\cdot | \mbmu, \mbSigma)$; where $\mbmu$ is the $J \times 1$ mean function vector and $\mbSigma$ is a $J \times J$ co-variance matrix composed by the pre-defined kernel function $\mathcal{K}(\cdot)$ over the inputs $\{\mbtheta_j\}_{j=1}^J$. In this paper, we adopt a standard radial basis function (RBF) kernel (see~\cite{brochu2010tutorial} for an overview of GP and the kernel functions). 

\begin{algorithm}[!ht]
    \caption{Pseudo-code of the workflow as a component of BO}
    \label{Alg1}
    \tcp{Initialization}
    Prepare MRI data $\mbOmega$ with $N$ patients and $M$ modalities, perform data filtering with quantile threshold $\gamma$\\
    \tcp{FAE training follows Figure~\ref{fig: flowchart}(B)}
    Compose $W$ pairs of modalities $G_{w=1}^W$, where $W = \binom{M}{2}$ \\
    Train FAE on $\{\mb{X}_m\}_{m=1}^M$ to generate latent features $\{\mb{Z_{m^{\prime}}}\}_{m^{\prime}=1}^W$ \\
    \tcp{Stability loss calculation follows Figure~\ref{fig: flowchart}(C)}
    \For {k =1,2,...,K} 
    {Randomly divide $\mbOmega$ into train ($\mbOmega_{train}$) and validation ($\mbOmega_{val}$) sets\\
    Produce latent pairs $\{\mb{Z}_{train,k}, \mb{Z}_{val,k}\}_{k=1}^K$\\
    \tcp{Pixel-wise clustering}
    Obtain $C_k$ and $C^{\prime}_k$ through standard K-means with $\eta$ clusters\\
    Compute $k$th stability loss $L_{s,k}$ by Eq~\eqref{Eq:stability}}
    Compute stability score $L_s$ by averaging over $\{L_{s,k}\}_{k=1}^K$ \\
    \tcp{Sub-region segmentation}
    Obtain patient-wise sub-region segments $\{\mb{P}_n\}_{n=1}^N$ \\
    \tcp{Patient-wise feature extraction}
    Extract $\{ \mb{F}_n\}_{n=1}^N$ for all $N$ patients\\
    \tcp{Survival analysis}
    Cluster patients into high/low risk subgroups based on $\{ \mb{F}_n \}_{n=1}^N$ using a standard K-Medoids algorithm. Perform survival analysis and obtain $p$ \\
    \tcp{BO loss calculation}
    Compute clinical significance score $L_p$ by Eq~\eqref{Eq:ploss}\\
    Compute joint loss $L$ follows Eq~\eqref{loss}
\end{algorithm}

Given training data $\mbOmega_{B} = \{\mbtheta_j, \mathcal{L}_j \}_{j=1}^J$, BO introduces a so-called acquisition function $a(\cdot)$ to propose the most probable candidate to be evaluated at each step. Amongst various types of acquisition functions~\cite{snoek2012practical}, we employ an EI strategy that seeks new candidates to maximize \textit{expected improvement} over the current best sample. Specifically, suppose $f'$ returns the best value so far, EI searches for a new $\mbtheta$ candidate that maximizes function $g(\mbtheta) = \max \{0, f'-f(\mbtheta)\}$. The EI acquisition can thus be written as a function of $\mbtheta$:
\begin{align}
a_{EI}(\mbtheta) = \mathbbm{E}( g(\mbtheta) | \mbOmega_{B})  
                   = \footnotesize (f' - \mbmu) \Phi(f' | \mbmu, \mbSigma) + \mbSigma \calN(f' | \mbmu, \mbSigma)
\label{EI}
\end{align}
where $\Phi(\cdot)$ denotes CDF of the standard normal distribution. In practice, BO step $J$ increases over time and the optimal $\theta_{\ast}$ can be obtained if the predefined convergence criteria is satisfied. Pseudo-code of the entire process is shown in both Algorithm~\ref{Alg1} and Algorithm~\ref{Alg2}.

\subsection{Experiment details}

Data from a total of $N = 117$ glioblastoma patients were collected and divided into training set $\mbOmega = 82$ and test set $\mbOmega_{test} = 35$, where the test set was separated for out-of-sample model evaluation. We collected both pMRI and dMRI data and co-registered them into T1C images, containing approximately 11 million pixels per modality over all patients. $M = 3$ input modalities were calculated, including rCBV (denoted as $\mb{r}$) from pMRI, and isotropic/anisotropic components (denoted as $\mb{p}$/$\mb{q}$) of dMRI, thus $\bf{X}= \{ \mb{p}, \mb{q}, \mb{r} \}$. Dataset $\mbOmega$ was used for stability loss calculation with $\mbOmega_{train} = 57$, $\mbOmega_{val} = 25$. $L_s$ was evaluated over $K=10$ trials for all following experiments. The BO is initialized with $J=10$ data points $\mbOmega_{B}$, $\gamma \in [0,1]$ and $\eta$ is an integer ranges between 3 and 7. The models were developed on Pytorch platform~\cite{paszke2019pytorch} under Python $3.8$. Both encoder $E$ and decoder $D$ employed a fully connected feed-forward NN with one hidden layer, where the hidden node number was set to $10$. We adopted \textit{hyperbolic tangent} as the activation function for all layers, \textit{mean squared error (MSE)} as the loss function, and \textit{Adam} as the optimiser. 


\section{Results and discussions} 
\label{sec:majhead}
We first present the clustering stability of the models incorporating FAE architecture, other AE variants against the baseline model and then compare the performance of the proposed methodology under different experimental settings. We finally demonstrate the results of survival analysis and independent test. 

\subsection{Evaluation of FAE based clustering}

The results comparing the models are detailed in Table~\ref{table stability}. One sees that all three AE variants show better stability performance than that of the baseline model in the varying cluster numbers. Of note, our proposed FAE architecture, which incorporates both standard AE and ensemble AE, outperforms other models in majority comparisons.

\begin{table}[!ht]
\caption{Stability performance of cluster algorithms under different AE variants. Baseline represents the original model without AE. The standard AE represents a standard 3-layer (with 1 hidden layer) feed-forward network and the ensemble AE is the FAE without dual decoder $D_s$. The hidden layer contains 10 nodes for all AE variants.}
\label{tab:similarity}
\resizebox{\columnwidth}{!}{%
\begin{tabular}{ccccc}
\hline
Clusters & 3 & 4  & 5 & 6 \\ \hline
\multicolumn{5}{c}{\textbf{Stability score}} \\ \hline
Baseline & 0.761±0.026 & 0.890±0.04 & 0.744±0.027 & 0.761±0.035\\
Standard AE &  0.909±0.024&0.896±0.063&0.859±0.06&0.836±0.061 \\
Ensemble AE & \textbf{0.972±0.013}&0.921±0.028&0.872±0.046&0.881±0.046\\
FAE & 0.909±0.048&\textbf{0.923±0.029}&\textbf{0.911±0.038}&\textbf{0.891±0.048} \\ \hline
\multicolumn{5}{c}{\textbf{Calinski-Harabasz (CH) score}} \\ \hline
Baseline  ($10^6$)& 4.12±0.00003 & 5.16±0.00013 & 4.82±0.00003 & 4.73±0.00009\\
Standard AE ($10^6$) &  5.94±0.63& 5.74±0.51 &5.50±0.41 & 5.36±0.28 \\
Ensemble AE  ($10^6$)& 10.43±0.67 & 10.99±0.52 & 10.98±0.89 & 11.09±1.00\\
FAE  ($10^6$)& \textbf{13.85±4.45} & \textbf{14.85±4.49} & \textbf{15.09±4.19} & \textbf{15.34±4.14} \\ \hline
\end{tabular}%
}\label{table stability}
\vspace{-0.4cm}
\end{table}

As expected, all AE variants enhance the clustering stability and quality, shown by the stability score and CH score. The latter of which is relatively sensitive to data scale but can provide reasonable evaluation for a fixed dataset. In our case, as the dimensions of the original input modalities and the latent features remain identical  ($M=3$), the considerably improved stability of the models incorporating FAE architecture suggests the usefulness of the FAE in extracting robust features for the unsupervised clustering.  Additionally, our experiments show that the FAE demonstrates remarkably stable performance in the clustering when the training data is randomly selected, which further supports the resilience of the FAE in extracting generalizable features for distance-based clustering algorithms.


\subsection{Adaptive hyper-parameter tuning}

\begin{figure}[!ht]
\centering
\subfigure[$\alpha = 0$ ]{
\includegraphics[width=5.9 cm]{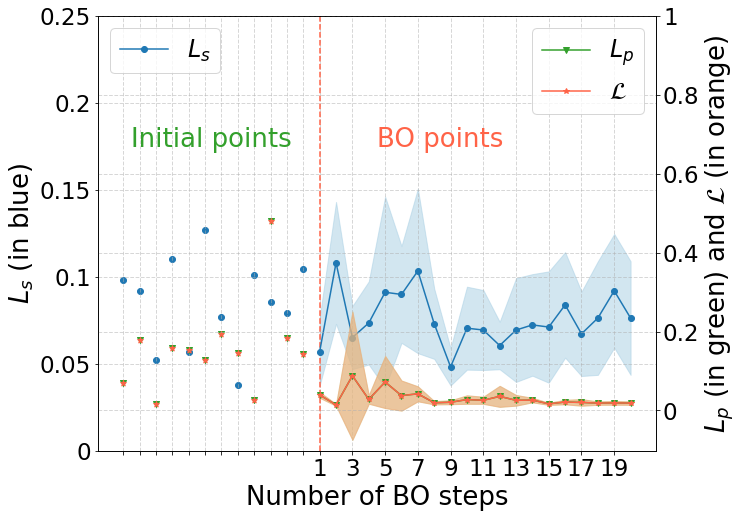}}
\subfigure[$\alpha = 0.25 $]{
\includegraphics[width=5.9 cm]{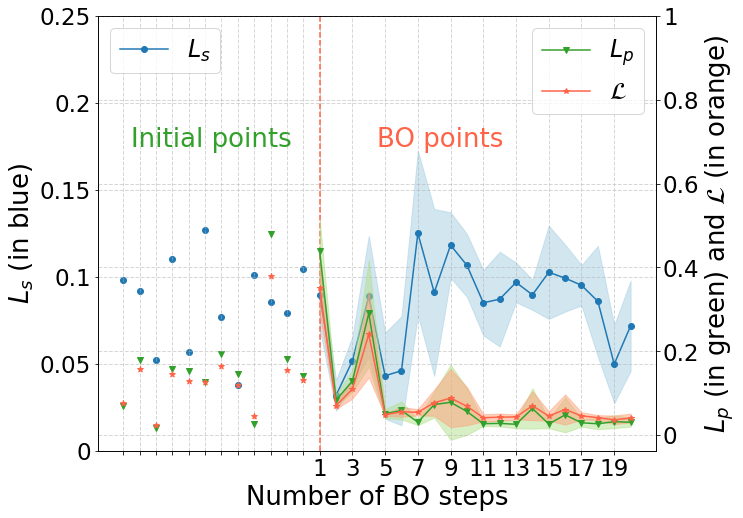}} \\
\subfigure[$\alpha = 0.5 $]{
\includegraphics[width=5.9 cm]{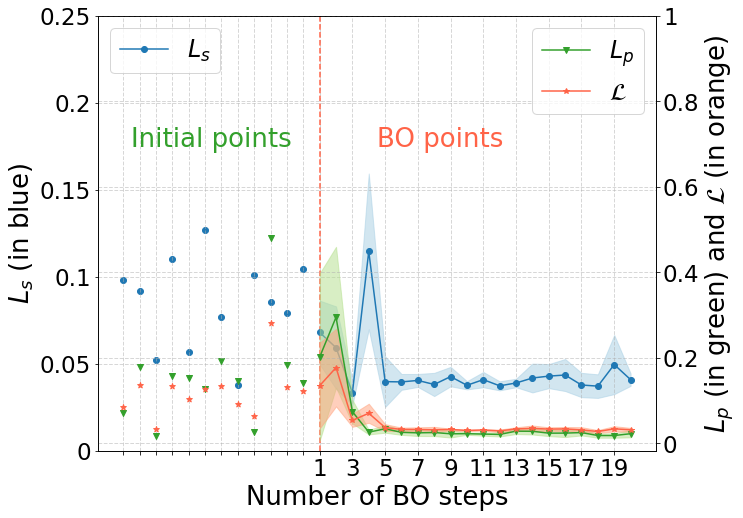}}
\subfigure[$\alpha = 1$ ]{
\includegraphics[width=5.9 cm]{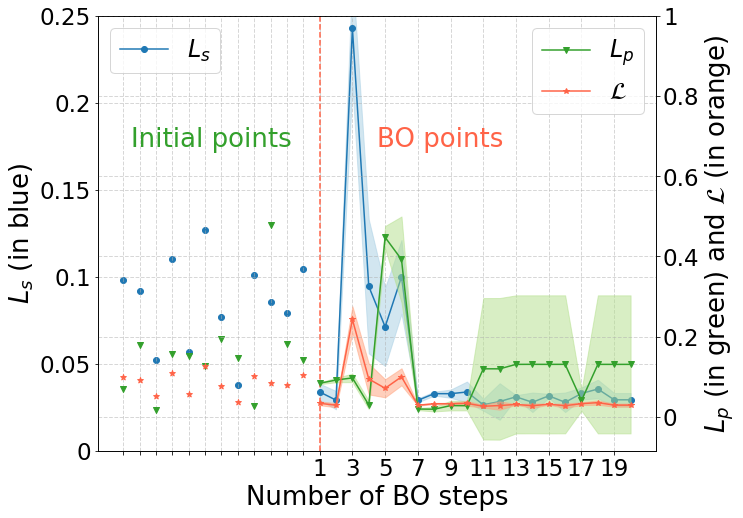}}
\caption{Performance of the proposed approach with respect to BO step number (on x-axis). Each figure contains two y-axis: stability loss $L_s$ (in blue) on the left y-axis, and both significant loss $L_p$ (in green) and joint loss (in orange) on the right y-axis. All losses are normalized and the shadowed areas in different colors indicate error-bars of the corresponding curves. Figure (a) - (d) shows the performance with loss coefficient $\alpha = 0, 0.25, 0.5$ and $1$, respectively.}
\label{fig:BOresults}
\vspace{-0.4cm}
\end{figure}

Figure~\ref{fig:BOresults} shows the performance of the proposed approach in 4 different $\alpha$ values in terms of stability score (lower score value indicates better stability). 10 initial training steps and 20 follow-up BO steps are evaluated in the experiments, all the results are averaged over 10 repeated trials. One sees significant dispersion of initial points (dots in the left half of each figure) in all figures, indicating reasonable randomness of initial points in BO training. BO proposes a new candidate $\mbtheta$ per step after the initial training. One observes that the joint loss $\mathcal{L}$ (orange curves) converges and the proposed approach successfully estimates (sub)optimal $\mbtheta_{\ast}$ in all $\alpha$ cases. 

Figure~\ref{fig:BOresults}(a) shows $\alpha=0$ case, for which $\mathcal{L} = L_p$ according to Equation~\eqref{loss}. In other words, the algorithm aims to optimize significance loss $L_p$ (green curve) rather than stability loss $L_s$ (blue curve). As a result, the orange and green curves overlap with each other, and the stability scores are clearly lower than that of $L_s$. A consistent trend can be observed across all four cases that the error-bar areas of $L_s$ (blue shadowed areas) shrink as the weight of $L_s$ increases in the joint loss. Similar observations can be seen in Figure~\ref{fig:BOresults}(d) where $\alpha=1$ and $\mathcal{L} = L_s$, the error-bar area of $L_p$ (green shadowed area) is considerably bigger than those in the rest $\alpha$ cases. Note that $L_s$ and $\mathcal{L}$ also overlap with each other and the mismatch in the figure is caused by the differences of left and right y-axis scale. When $\alpha=0.5$ (Figure~\ref{fig:BOresults}(c)), clustering stability can quickly converge in a few BO steps (around 6 steps in the orange curve), shows the advantage of the proposed BO integrated method in hyper-parameter optimization.  

\subsection{Statistical analysis and independent test}

\begin{figure}[!ht]
\centering
\subfigure[Train set $\mbOmega = 82$ patients]{
\includegraphics[width=5.9 cm]{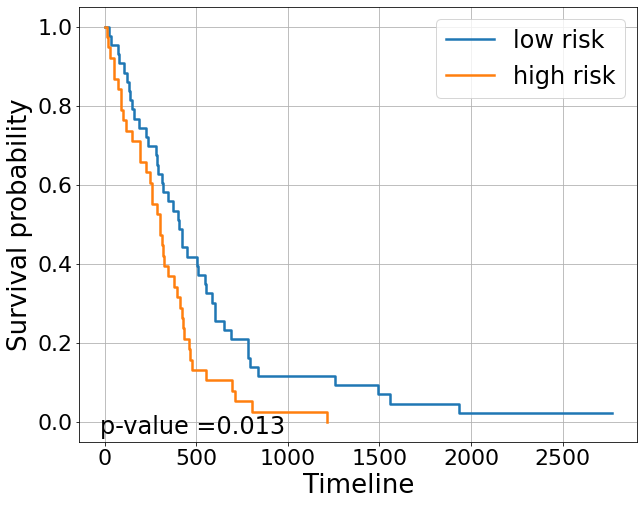}}
\subfigure[Test set $\mbOmega_{test} = 35$ patients]{
\includegraphics[width=5.9 cm]{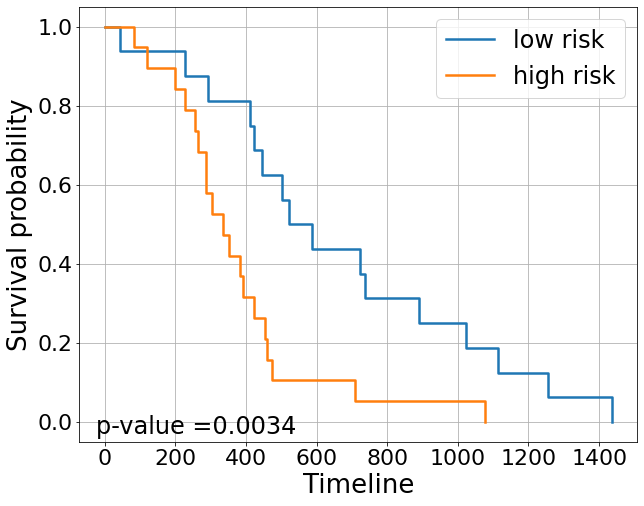}}
\caption{KM survival curves for the train and test datasets.}
\label{fig:survival}
\vspace{-0.4cm}
\end{figure}

\begin{figure}[!ht]
\centering
\subfigure[low-risk (CE)]{
\includegraphics[width=2.5cm]{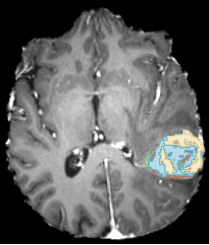}}
\subfigure[low-risk (NE)]{
\includegraphics[width=2.5cm]{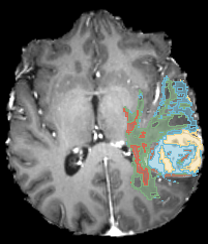}}
\subfigure[high-risk (CE)]{
\includegraphics[width=2.5cm]{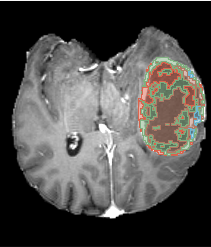}}
\subfigure[high-risk (NE)]{
\includegraphics[width=2.5cm]{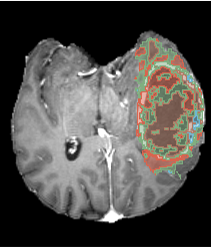}}

\
\caption{Two case examples from the high-risk  (a \& b) and lower-risk  (c \& d) group, respectively. Different colours denote the partitioned sub-regions. The two patients have significantly different proportions of sub-regions with clinical relevance, which could provide clinical decision support. }
\label{fig:tumor2}
\vspace{-0.4cm}
\end{figure}

Upon convergence of BO, we acquire well-trained FAE encoders to extract features from modalities, a well-trained clustering model for tumor sub-region segmentation and a population-level grouping model to divide patients into high-risk and low-risk subgroups. Subsequently, we apply these well-trained models to the test set with 35 patients. The results of KM analysis are shown in Figure ~\ref{fig:survival}, illustrating that the spatial features extracted from tumor sub-regions could lead to patient-level clustering that successfully separates patients into distinct survival groups in both datasets (Train: p-value = 0.013 Test: p-value = 0.0034). Figure~\ref{fig:tumor2} shows two case examples from the high-risk and low-risk subgroups, respectively, where different colours indicate the partitioned sub-regions. Intuitively, these sub-regions are in line with the prior knowledge of proliferating, necrotic, and edema tumor areas, respectively. 

\section{Conclusions}
\label{sec:print}

The paper is an interdisciplinary work that helps clinical research to acquire robust and effective sub-regions of glioblastoma for clinical decision support. The proposed FAE architectures significantly enhance the robustness of the clustering model and improve the quality of clustering results. Additionally, robust and reliable clustering solutions can be accomplished with minimal time investment by integrating the entire process inside a BO framework and presenting a unique loss function for problem-specific multi-task optimization. Finally, the independent validation of our methodology using a different dataset strengthens its viability in clinical applications.

Although we have conducted numerous repeating trials, it is inevitable to eliminate the randomness for clustering algorithm experiments. In future work, we could include more modalities and datasets to test the framework. To enhance the clinical relevance, more clinical variables could be included into the BO framework for multi-task optimization. To summarise, the BO framework combined with the suggested FAE and mixed loss represents a robust framework for obtaining  clustering results that are clinically relevant and generalizable across datasets.




\bibliographystyle{splncs04.bst}
\bibliography{refs.bib}
\end{document}